\documentclass[journal]{IEEEtran}
\usepackage{graphicx}
\usepackage{graphics}
\usepackage{caption}
\usepackage{lipsum}

\usepackage[T1]{fontenc}
\usepackage[utf8]{inputenc}
\usepackage[british,UKenglish,USenglish,english,american]{babel}
\usepackage{courier}
\usepackage{authblk}
\usepackage{subcaption}
\usepackage[square,numbers]{natbib}
\setcitestyle{numbers}

\usepackage{booktabs,siunitx,array,threeparttable, makecell}
\usepackage{makecell}
\usepackage{amsmath}
\usepackage{float}
\usepackage{listings}
\usepackage{alltt} 
\usepackage{natbib}
\usepackage{pifont}
\newcommand{\cmark}{\ding{51}}%
\newcommand{\xmark}{\ding{55}}%
\ifCLASSINFOpdf
\else
\fi
\begin{document}
%
\title{Quantized Transformer Language Model Implementations on Edge Devices}
%
%
%
\author{\IEEEauthorblockN{Mohammad Wali Ur Rahman\IEEEauthorrefmark{1}, 
Murad Mehrab Abrar\IEEEauthorrefmark{1}, Hunter Gibbons Copening\IEEEauthorrefmark{2}, Salim Hariri\IEEEauthorrefmark{1}, Sicong Shao\IEEEauthorrefmark{4}, Pratik Satam\IEEEauthorrefmark{3}, and Soheil Salehi\IEEEauthorrefmark{1}}\\
\IEEEauthorblockA{\IEEEauthorrefmark{1}Electrical and Computer Engineering, University of Arizona, Tucson, AZ 85721 USA\\
\IEEEauthorrefmark{2}Computer Science, University of Arizona, Tucson, AZ 85721 USA\\
\IEEEauthorrefmark{3}Systems and Industrial Engineering, University of Arizona, Tucson, AZ 85721 USA\\
\IEEEauthorrefmark{4}Electrical Engineering and Computer Science, University of North Dakota, Grand Forks, ND 58202 USA\\
Email: \{\IEEEauthorrefmark{1}mwrahman, \IEEEauthorrefmark{1}abrar, \IEEEauthorrefmark{2}huntercopening, \IEEEauthorrefmark{1}hariri, \IEEEauthorrefmark{3}pratiksatam, \IEEEauthorrefmark{1}ssalehi\}@arizona.edu; \IEEEauthorrefmark{4}sicong.shao@und.edu}}
\vspace{-0.15in}
%
%

\markboth{22\textsuperscript{nd} International Conference of Machine Learning and Applications, ICMLA 2023}%
{Rahman \MakeLowercase{\textit{et al.}}: Quantized Transformer Language Model Implementations on Edge Devices}
%



\maketitle

\begin{abstract}
  Large-scale transformer-based models like the Bidirectional Encoder Representations from Transformers (BERT) are widely used for Natural Language Processing (NLP) applications, wherein these models are initially pre-trained with a large corpus with millions of parameters and then fine-tuned for a downstream NLP task. One of the major limitations of these large-scale models is that they cannot be deployed on resource-constrained devices due to their large model size and increased inference latency. In order to overcome these limitations, such large-scale models can be converted to an optimized FlatBuffer format, tailored for deployment on resource-constrained edge devices. Herein, we evaluate the performance of such FlatBuffer transformed MobileBERT models on three different edge devices, fine-tuned for Reputation analysis of English language tweets in the RepLab 2013 dataset. In addition, this study encompassed an evaluation of the deployed models, wherein their latency, performance, and resource efficiency were meticulously assessed. Our experiment results show that, compared to the original BERT large model, the converted and quantized MobileBERT models have 160$\times$ smaller footprints for a 4.1\% drop in accuracy while analyzing at least one tweet per second on edge devices. Furthermore, our study highlights the privacy-preserving aspect of TinyML systems as all data is processed locally within a serverless environment.
\end{abstract}

\begin{IEEEkeywords}
IoT, Natural Language Processing, Machine Learning, BERT, Reputation Polarity, Social Media, Embedded Systems, TinyML, Privacy.
\end{IEEEkeywords}

%
\IEEEpeerreviewmaketitle

\section{Introduction}
Pre-trained large-scale Natural Language Processing (NLP) models have been exhibiting remarkable performance in most NLP tasks using transformer-based architectures. By stacking multiple encoder/decoder layers, combined with attention mechanism \cite{vaswani2017attention}, these architectures are producing promising results in the field of NLP. Models such as BERT \cite{devlin2018bert}, RoBERTa \cite{liu2019roberta}, XLNet \cite{yang2019xlnet}, and GPT-4 \cite{openai2023gpt4} have been increasingly popular in the commercial development of various smart AI systems to analyze audio/text input. These services will be integrated into mobile computing and Internet of Things (IoT) devices to improve user experience, making it imperative to deploy such NLP services on resource-constrained edge devices to improve the service response times \cite{niu2020real}. 

However, these transformer-based NLP models are pre-trained using TensorFlow (or similar) end-to-end machine learning platform and they are optimized for classification accuracy, making them contain thousands of layers of neurons with a large number of optimization parameters. Such models are large in size and require significant memory for storage and processing. \cite{devlin2018bert, liu2019roberta, yang2019xlnet, radford2019language}. Accommodation of such large-scale models in edge devices with smaller storage is a major challenge. In addition to the storage needs, the latency, and the computational cost also prove to be huge obstacles to the deployment of traditional machine learning (ML) models \cite{wang2020hat}. Due to the increased latency resulting from the constrained resources of edge devices, the conventional deep learning models often fail to meet the real-time requirements \cite{niu2020real}. To address these challenges, TinyML has proven to be a promising candidate. 

One of the main focuses of the field of TinyML is on developing and deploying ML models on resource-constrained, small, and low-power devices such as microcontrollers, sensors, and edge devices \cite{warden2019tinyml}. Traditionally IoT device services rely on sending data to a remote server for ML analysis (to provide services), adding performance delays, increasing the service's dependence on the availability and quality of the communication network, and posing security challenges, including those concerning user privacy \cite{marin2022serverless}. Integration of TinyML-based models into the service allows the deployment of these ML-based services on the device itself or an edge node, mitigating the aforementioned challenges. TinyML allows for real-time data processing at the edge, enabling a more efficient and faster decision-making process, which can be crucial for some applications, such as autonomous systems, robotics, and industrial automation \cite{soro2021tinyubq}.  

TensorFlow-Lite \cite{TensorFlowLite} is an example of a TinyML-based algorithm that is optimized for deployment on embedded devices \cite{abadi2016tensorflow}. It includes a number of features that make it well-suited for implementing TinyML, such as support for on-device ML, quantization and pruning of models to reduce their size and improve performance, and a small footprint that allows it to run on devices with limited memory and storage. To the best of our knowledge, no previous studies have thoroughly analyzed the performance and resource requirements of smaller BERT variants such as MobileBERT \cite{sun2020MobileBERT} on resource-constrained devices. Our research aims to provide insight into the potential capabilities and limitations of utilizing MobileBERT on Raspberry Pi devices for NLP tasks. 
The contributions of this paper are as follows:
\begin{itemize}
\item Provide a comparative performance and resource usage analysis of BERT Large and its lightweight variant, MobileBERT.
\item Develop a novel framework for evaluating MobileBERT models in TensorFlow-Lite format on resource-constrained devices, both with and without quantization applied.
\item Implement a low-cost and privacy-preserving system that processes data locally on edge devices using a TinyML model.
\item Demonstrate the performance of MobileBERT in classifying social media texts based on their reputation polarity.
\end{itemize}



\section{Related Works}
TinyML has seen significant growth in recent years, with many research studies addressing the challenges and opportunities associated with deploying ML models on small, low-power devices. In this section, we will review the most relevant literature on TinyML, focusing on recent advancements and challenges in the field. Table \ref{tab:related-works} compares the related works qualitatively.
There have been many studies with the aim of compression of large neural networks. Previous studies have highlighted the importance of model compression techniques for deploying ML models on resource-constrained devices. In \cite{han2015deep}, Han et al. present a technique for compressing deep neural networks by using three methods, pruning, trained quantization, and Huffman coding, to reduce the size and computational cost of the model while maintaining good accuracy. Iandola et al. in \cite{iandola2016squeezenet} present a new Convolutional Neural Network (CNN) architecture that achieves AlexNet-level accuracy while having 50$\times$ fewer parameters and a model size of less than 0.5MB. Additionally, in \cite{jacob2018quantization}, Jacob et al. first showed that quantizing neural networks to perform inference using integer-only arithmetic without a significant loss in accuracy is possible. The paper proposes a training method for quantization that combines quantization-aware training and post-training quantization. 

Moreover, Wang et al. present a method for automatically quantizing deep learning models for efficient deployment on hardware devices such as mobile phones, embedded systems, and IoT edge devices. Their proposed method, named Hardware-Aware Automated Quantization (HAQ), uses reinforcement learning (RL) and evolution algorithms to explore the quantization search space and find the best quantization method for a given hardware target \cite{wang2019haq}. Additionally, the AMC method proposed in \cite{he2018amc} uses a hardware-aware evaluation function and a resource constraint such as FLOPs to control the search space and find the best trade-off between model size and performance. It also uses an RL algorithm to search for the best combination of model compression techniques and hyperparameters that maximize the trade-off between model performance and resource efficiency.

\begin{table}[!t]
    \centering
    \caption{Qualitative Comparison of the Related Works}
    \label{tab:related-works}
    \resizebox{\columnwidth}{!}{%
    \begin{tabular}{cccccc}
    \hline \hline
       \thead{Proposed \\ System} & \thead{Experiment \\ Resources} & \thead{Transformer-based \\ Large Architectures} & \thead{CPU Utilization \\ Details Presented} & \thead{Memory Utilization \\  Details Presented} & \thead{Power Dissipation\\ Details Presented}\\ \hline
       \thead{Deep \\ Compression \cite{han2015deep}} & \thead{Intel Core i7 5930K \\ NVIDIA GeForce GTX \\ Titan X, NVIDIA Tegra} & \color{red}\Large\xmark\color{black} & \color{green}\Large\cmark\color{black} & \color{red}\Large\xmark\color{black} & \color{green}\Large\cmark\color{black} \\ 
       AMC \cite{he2018amc} & \thead{Qualcomm Snapdragon \\ 821, NVIDIA Titan \\ XP GPU} & \color{red}\Large\xmark\color{black} & \color{green}\Large\cmark\color{black} & \color{red}\Large\xmark\color{black} & \color{red}\Large\xmark\color{black}\\
       \thead{Houlsby et al. \cite{houlsby2019parameter}} & \thead{Google Cloud \\ TPU} & \color{green}\Large\cmark\color{black} & \color{red}\Large\xmark\color{black} & \color{red}\Large\xmark\color{black} & \color{red}\Large\xmark\color{black}\\ 
       \thead{Lite \\ Transformers \cite{wu2020lite}} & \thead{ARM Cortex-A72 \\ mobile CPU} & \color{green}\Large\cmark\color{black} & \color{red}\Large\xmark\color{black} & \color{red}\Large\xmark\color{black} & \color{red}\Large\xmark\color{black} \\
       HAT \cite{wang2020hat} & \thead{Intel Xeon, NVIDIA Titan, \\ ARM Cortex A72} & \color{green}\Large\cmark\color{black} & \color{red}\Large\xmark\color{black} & \color{red}\Large\xmark\color{black} & \color{red}\Large\xmark\color{black}\\
       \thead{Niu et al. \cite{niu2021compression}} & \thead{Qualcomm \\ Snapdragon 865} & \color{green}\Large\cmark\color{black} & \color{green}\Large\cmark\color{black} & \color{red}\Large\xmark\color{black} & \color{red}\Large\xmark\color{black}\\
       This Work & \thead{Intel Core i5-7500 \\ Broadcom BCM2837 1.2 Ghz \\ Broadcom BCM2837 1.4 Ghz \\ Broadcom BCM2711 1.5 Ghz} & \color{green}\Large\cmark\color{black} & \color{green}\Large\cmark\color{black} & \color{green}\Large\cmark\color{black} & \color{green}\Large\cmark\color{black}\\
    \hline \hline
    \end{tabular}
}\vspace{-0.15in}
\end{table}

\begin{figure*}[!t]
    \centering
    \captionsetup{justification=centering}
    \includegraphics[width=0.9\textwidth]{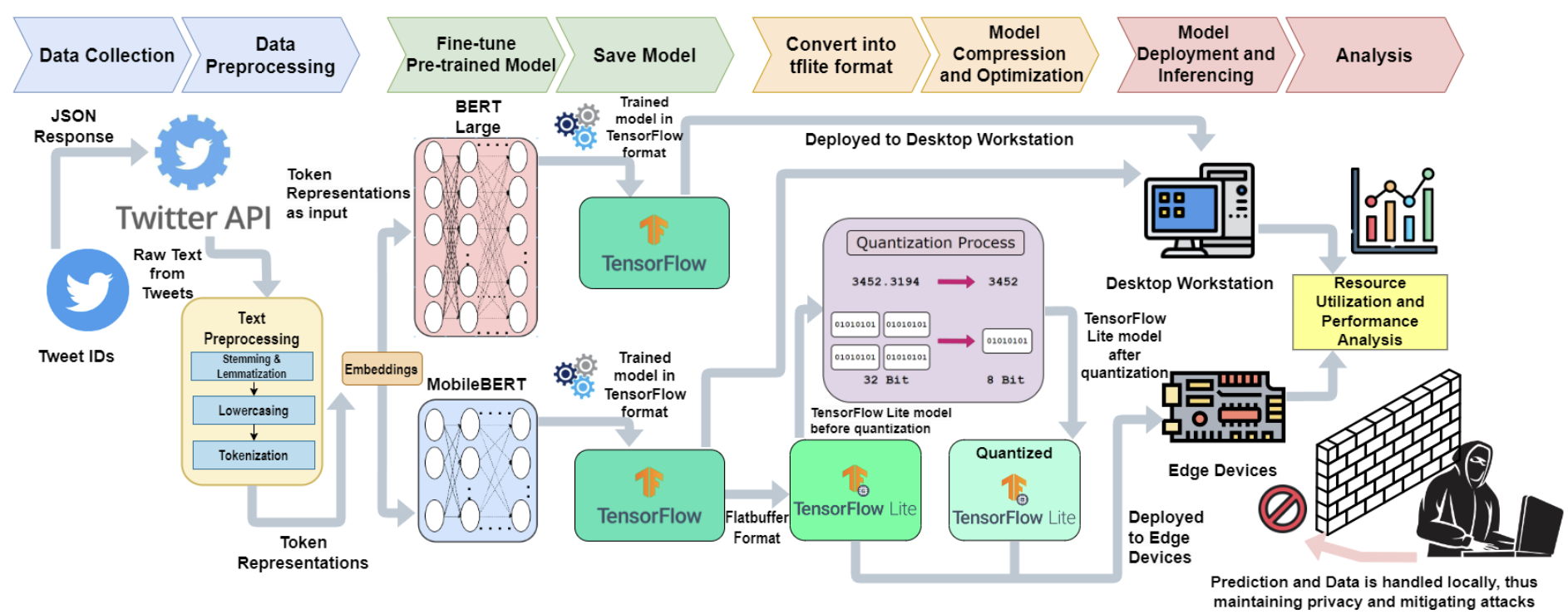}
    \vspace{-0.5em}
    \caption{Proposed Framework Architecture}
    \label{fig:galaxy}
    \vspace{-2em}
\end{figure*}

Furthermore, there have been some recent advancements of TinyML in the NLP area as well, where the smaller NLP models have been able to achieve comparable performance in contrast with full-precision models while using significantly fewer parameters. Houlsby et al. in \cite{houlsby2019parameter} present an adapter-dependent method for transfer learning in NLP tasks that is efficient in terms of the number of parameters used, where only the higher layers are fine-tuned for the specific task at hand to achieve parameter efficiency, as they are useful in building task-specific features. In \cite{wu2020lite}, Wu et al. present a novel transformer architecture called LITE Transformer, where the authors introduced a long-short-range attention mechanism, which selectively attends to different ranges of positions in the input sequence based on their relevance to the task. This reduces the number of attention calculations required, resulting in a more efficient model. In the work presented in \cite{yan2020micronet}, Yan et al. proposed a transformer-based architecture called Micronet, a parameter and computation-efficient language model. The architecture is based on a combination of techniques such as adaptive embedding, knowledge distillation, network pruning, low-bit quantization, and differentiable non-parametric cache. The approach performs similarly to other full-precision models in various NLP tasks with fewer parameters. 

In \cite{wang2020hat}, the authors propose an efficient and adaptive transformer architecture that takes into account the characteristics of hardware, such as memory bandwidth, computation power, and energy consumption. The proposed model uses a combination of techniques such as knowledge distillation, quantization, and model pruning to reduce computation and memory requirements. Both TinyBERT \cite{jiao2019tinybert} and MobileBERT \cite{sun2020MobileBERT} are compact and efficient versions of BERT that can run on resource-constrained devices, and both can be used in a wide range of applications, such as offline natural language understanding on mobile devices, voice assistants, and language-based IoT applications. Moreover, Niu et al. \cite{niu2021compression} deployed their own compiler-aware neural architecture optimization models in addition to other BERT variants, including MobileBERT in TensorFlow-Lite format, and compared the performances in Question-answering and Text Generation tasks. Their proposed framework, as well as the other architectures, were evaluated using the Samsung Galaxy S10 cell phone, which has a Qualcomm Snapdragon 865 processor. 

Many works have analyzed the performance and resource requirements of TensorFlow-Lite MobileBERT models on resource-constrained devices. Despite their valuable efforts, these works fail to comprehensively analyze their implementations' resource utilization. Herein, we provide a comprehensive assessment of the capabilities and limitations of MobileBERT on Raspberry Pi devices for NLP tasks, thereby filling the existing gap in the literature. In our study, for the specific NLP task under consideration, we fine-tuned both the BERT Large and MobileBERT models to perform multiclass classification based on the polarity of reputation, similar to the approach presented in \cite{rahman2022bert}. Furthermore, we demonstrate the trade-off between performance and model size, as well as latency reduction, when utilizing quantized TensorFlow-Lite models. Our findings indicate that significant improvements in terms of model size and latency are achieved while incurring a negligible decrease in performance. 

\section{Framework Architecture}
In this section, we present the architecture of our framework. The overall architecture of our proposed framework is demonstrated in Figure \ref{fig:galaxy}, which illustrates the various stages of the framework. The first stage of the framework involves data collection and preprocessing, which serves as an essential step for fine-tuning the BERT large and MobileBERT models. Subsequently, the fine-tuning phase is carried out using the preprocessed data. After the training process is completed, the MobileBERT model is converted to the TensorFlow-Lite model format, with subsequent optimization and compression achieved through the application of quantization. The analysis module then evaluates the performance and resource utilization of the models deployed on the target machines. Throughout the system, all processing occurs on the local edge device and does not require communication with a central server, reducing the risk of data interception and ensuring data privacy. 

\vspace{-0.1in}
\subsection{Data Collection and Preprocessing}
Herein, our primary objective was to evaluate the performance of TensorFlow Lite models compared to TensorFlow models on edge devices. To accomplish this, first, we conducted fine-tuning of the BERT Large and MobileBERT models on a supervised dataset \cite{amigo2013overview} that was specifically curated for reputation research. The multiclass classification task based on reputation polarity served as a means to assess and compare the performance of the TensorFlow Lite models within the context of our experiments. To accomplish this, we have utilized the RepLab 2013 \cite{amigo2013overview} dataset, which comprises tweets about 61 entities from 4 different domains. For the fine-tuning process, we have focused exclusively on the English tweets within the dataset, as MobileBERT is not optimized for multilingual tasks. 

In this study, we used the Twitter API \cite{twitterapi} to collect tweets for our analysis. The Twitter API is a powerful tool that allows developers to access a wide range of data from the Twitter platform, such as tweets and their associated metadata. The Twitter API returns JSON objects containing the tweets that match our search criteria. Then we proceed to pre-process the extracted texts. Text pre-processing is a vital stage in reputation polarity tasks as it converts social media text into a more consumable format that is more suitable for ML models. Through this process, tweets from the RepLab 2013 dataset are cleaned and prepared for model training. In particular, we remove redundant spaces, symbols, emojis, URL links, and punctuation marks to ensure the data is in a compatible format for the ML model. The texts afterward are tokenized using the appropriate BERT tokenizer for the BERT models.

\vspace{-0.1in}
\subsection{Fine-tuning of the Pre-trained Models}
BERT \cite{devlin2018bert} is a significant innovation in contextualized representation learning for NLP. In their work \cite{devlin2018bert}, authors demonstrate that even though the word embedding layer in traditional deep learning models for NLP tasks is trained on large language corpora, training a range of neural network architectures that encode contextual representations solely based on the limited supervised data for end NLP tasks is still inadequate. BERT employs a fine-tuning process that requires minimal architecture modifications for each end NLP task. BERT offers two parameter-intensive configurations, BERT Base and BERT Large. BERT Base comprises 768 hidden dimensions, 12 transformer blocks, and 12 attention heads, with a total of 110 million parameters. BERT Large, on the other hand, has 1024 hidden dimensions, 24 transformer blocks, and 16 attention heads, totaling 340 million parameters. The pre-training process with BERT models involves two key methods: masked language modeling and next-sentence prediction.
In order to provide a comprehensive assessment of the capabilities of MobileBERT, we have selected BERT large as the baseline model for comparison in our study. This decision is based on the fact that MobileBERT was derived from the inverted bottleneck version of BERT Large (IB-BERT) \cite{sun2020MobileBERT} through the process of knowledge distillation. In essence, MobileBERT represents a lightweight version of IB-BERT, specifically optimized for use on resource-constrained edge devices. This comparison allows us to evaluate the trade-offs between the performance and resource demands of MobileBERT and BERT large and to provide insight into the potential of MobileBERT for NLP tasks on edge devices.

Both BERT large and MobileBERT models use token representation vectors as input during the fine-tuning process. Each token is represented by a sum of three representation vectors: a positional embedding vector, which encodes information about the token's location in the sequence; a sentence vector is used when a single sentence is not sufficient to convey the context.; and a typical word embedding vector, which is a vector representation of the word in context. Additionally, BERT extends the input sentence by incorporating the [SEP] token and the [CLS] token. The [CLS] token carries the embedding for specific classification tasks, whereas the [SEP] token is responsible for separating segments. For our reputation polarity task, BERT utilizes the last hidden state \textit{h} derived from the initial token [CLS] to encapsulate the entire input sequence. To predict the probability of reputation polarity class $c$, we augment BERT with a softmax classifier positioned atop, employing 
\begin{equation}
p(c|h) = softmax(Vh), 
\end{equation}
where the parameter matrix \textit{V} corresponds to the reputation polarity prediction task. Through fine-tuning the reputation polarity training data, we simultaneously optimize all parameters of BERT and the parameter matrix \textit{V}. After completing the training process, the BERT Large TensorFlow model is deployed on a desktop workstation for further evaluation. Meanwhile, the MobileBERT model undergoes a conversion process to TensorFlow-Lite format, and compression techniques are applied to optimize it for deployment on resource-constrained edge devices.


\subsection{Model Compression and Optimization}

The trained TensorFlow MobileBERT models need to be converted to TensorFlow-Lite format. To achieve this, the first step is to export the TensorFlow model to a file format that TensorFlow-Lite can read, such as a TensorFlow SavedModel or a frozen TensorFlow GraphDef. This can be done using TensorFlow's built-in export functions. Freezing the model involves converting the variables in the model to constants so that the model's weights are embedded in the model graph. The TensorFlow-Lite Converter is then used to convert the frozen model to a TensorFlow-Lite FlatBuffer file. The converter takes the frozen model as input and generates a TensorFlow-Lite model. The converted TensorFlow-Lite models go through further quantization. We have used the Dynamic Range Quantization technique \cite{TensorFlowQuantization} as our quantization process. In DRQ, the range of the weights and activations are adjusted based on the data range and are converted from float points to 8-bit integers. This allows for more efficient use of the available bits while offering a smaller model size and lower computational requirements without a significant loss of performance. In contrast, in the traditional quantization method, the range of the weights and activations of the models are fixed, which can lead to information loss and a significant drop in performance. Both the quantized and non-quantized models are then deployed in target machines, and their performance and resource utilization data are analyzed.

\begin{figure}[!t]
    \centering
    \captionsetup{justification=centering}
    \includegraphics[width=0.95\columnwidth]{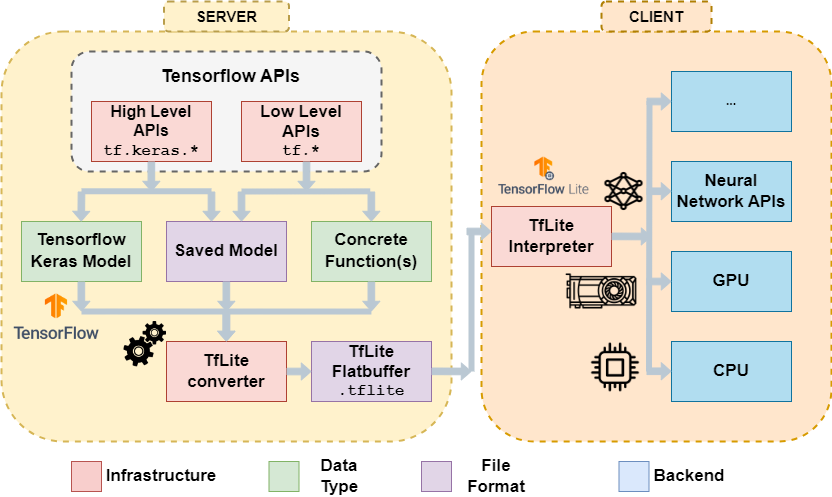}
    \caption{TensorFlow-Lite Conversion and Deployment.}
    \label{fig:tlite}
\end{figure}

\begin{table}[!t]
    \centering
    \caption{Hardware Specifications of Devices}
    \label{tab:hwspec}
    \vspace{-0.5em}
    \resizebox{0.75\columnwidth}{!}{%
    \begin{tabular}{cccc}
    \hline\hline
        Machine & CPU & RAM & Memory \\ \hline
        \thead{Embedded\\Raspberry Pi 3B} & \thead{Broadcom BCM2837 \\ SoC @1.2GHz} & 1GB & \thead{32GB \\ $\mu$SD} \\
        \thead{Embedded\\Raspberry Pi 3B+} & \thead{Broadcom BCM2837 \\ SoC @1.4GHz} & 1GB & \thead{32GB \\ $\mu$SD} \\
        \thead{Embedded\\Raspberry Pi 4B} & \thead{Broadcom BCM2711 \\ SoC @1.5GHz} & 4GB & \thead{32GB \\ $\mu$SD}\\ \hline\hline
    \end{tabular}
}
\vspace{-2em}
\end{table}

\subsection{Model Deployment}
To deploy the TensorFlow-Lite models on resource-constrained devices, we have utilized the TensorFlow-Lite (TFLite) interpreter \cite{tf.lite.Interpreter}. TFLite interpreter is a library that enables developers to run TensorFlow-Lite models on edge devices with limited computational resources. It takes a TensorFlow-Lite model as input and performs the computations defined in the model's graph by loading the model into memory and converting it into a format that can be executed on the device's hardware. It provides an API that enables developers to interact with the model, such as inputting data, running the computations, and retrieving the output. Thus, the TFLite interpreter bridges the gap between the TensorFlow-Lite model and the device's hardware. Figure \ref{fig:tlite} illustrates the process of converting TensorFlow models to TensorFlow-Lite format and the deployment of the converted models on target machines. In this study, the TensorFlow models were deployed on a desktop workstation, and the inference was performed solely using the CPU on the test dataset. Additionally, both quantized and non-quantized versions of the TensorFlow models were deployed on Linux-based embedded devices. Data pertaining to performance metrics and resource utilization obtained from the inference operation on these devices were collected and utilized for comparative analysis.

\subsection{Model Evaluation and Analysis}
To conduct a comprehensive evaluation of the deployed models, we introduce metrics that considered model latency, performance, and efficiency. These metrics, namely the Speed Index (SI), Model Performance Index (MPI), and Resource Efficiency Ratio (RER), were designed to provide a holistic assessment of the models' effectiveness.
\subsubsection{Speed Index (SI)}
The Speed Index (SI) metric captures the trade-off between the speed of the model, represented by FLOPS (Floating-Point Operations Per Second), quantization bits, and total time in seconds. SI metric is computed using the following equation:
\begin{equation}
SI = \frac{{{FLOPS}}}{{Q}\times{t}},
\end{equation}
where $Q$ is the Quantization Bits, and $t$ is the total time in \textit{seconds}. This metric quantifies how fast the model performs in relation to the number of operations, quantization, and time. 

\subsubsection{Model Performance Index (MPI)}

The Model Performance Index (MPI) metric evaluates the overall performance of the model by considering the average accuracy ($Accuracy_{avg}$), average F-Score ($(F-Score)_{avg}$), and total power dissipation ($Power_{tot}$) in \textit{Kilo Watts}. The MPI is computed using the following equation:
\begin{equation}
MPI = \frac{Accuracy_{avg} + {(F-Score)}_{avg}}{Power_{tot}},
\end{equation}
which quantifies the model's performance in terms of accuracy and energy efficiency. 

\subsubsection{Resource Efficiency Ratio (RER)}

The Resource Efficiency Ratio metric measures the efficiency of resource utilization by considering CPU utilization ($CPU\%$), memory utilization ($MEM\%$), and energy consumption ($Energy_{tot}$). The Resource Efficiency Ratio is computed as follows:
\begin{equation}
RER = \frac{Energy_{tot}}{CPU\% \times MEM\%},
\end{equation}
which quantifies how efficiently the model utilizes resources. 

\begin{table}[!t]
    \centering
    \caption{Experimental Results}
    \label{tab:results}
    \vspace{-0.5em}
    \resizebox{\columnwidth}{!}{%
    \begin{tabular}{cccccc}
    \hline\hline
    Model &Hardware &\thead{Avg. Accuracy} &\thead{Avg. F-Score} &\thead{Max Accuracy} &\thead{Max F-Score}\\\hline
    \thead{TensorFlow-Lite\\MobileBERT\\(32 bit)} &\thead{Embedded\\Device}  &0.685$\pm$0.0031 &0.602$\pm$0.004 &0.70 &0.61\\
    \thead{TensorFlow-Lite\\MobileBERT\\(16 bit)} &\thead{Embedded\\Device} &0.683$\pm$0.0179 &0.601$\pm$0.002 &0.69 &0.61\\
    \thead{TensorFlow-Lite\\MobileBERT\\(8 bit)} &\thead{Embedded\\Device} &0.684$\pm$0.0027 &0.603$\pm$0.004 &0.69 &0.61\\
    \hline\hline
    \end{tabular}
}\vspace{-1.5em}
\end{table}

\section{Experiment Results}
This Section describes the experimental settings and presents a comprehensive discussion of the results.

\subsection{Experiment Settings}

In this study, the performance of the models is evaluated using the Accuracy and F-score metrics, consistent with the evaluation methods employed in RepLab 2013 \cite{amigo2013overview}. In both models, the learning rate employed was $1\times{10^{-5}}$ and the batch size utilized was 32. To provide the reproducibility of results, the experiments were performed with five different random seeds on all devices during each iteration of the experiments. Additionally, resource utilization data were collected and monitored in three stages. 

\begin{figure}[!t]
\centerline{\includegraphics[width=0.8\columnwidth]{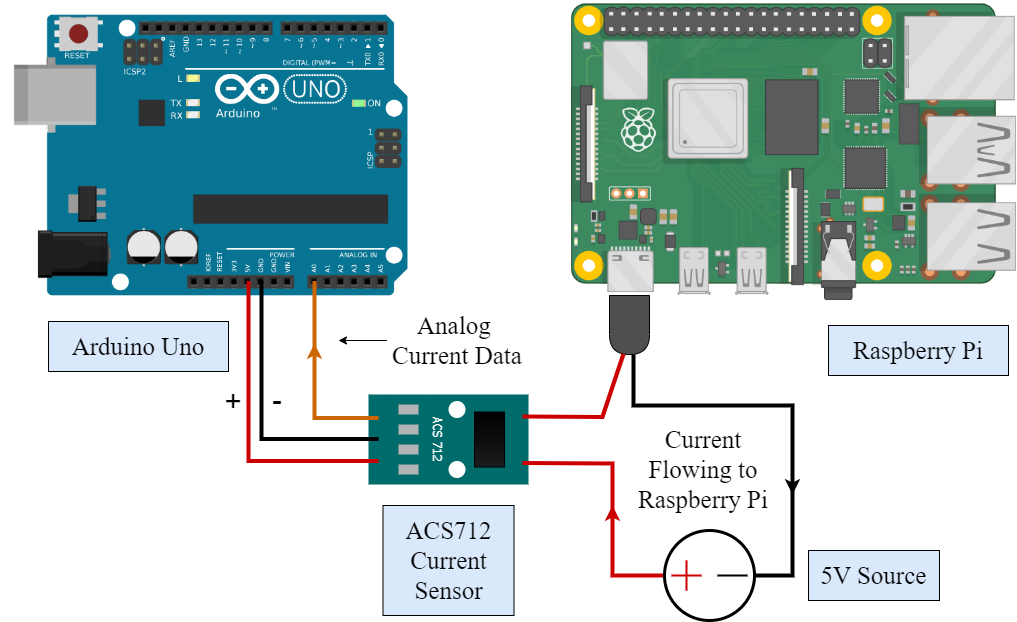}}  
\vspace{-0.5em}
\caption{Current Sensing Hardware Setup Diagram.}
\label{current sensing}
\vspace{-1.5em}
\end{figure}


\subsubsection{CPU and Memory Utilization Data using PSUTIL}
Prior to the initiation of the experiment, system-wide CPU and memory utilization percentage data were gathered. During the experiment, simultaneous collection of both system-wide and process-specific CPU and memory utilization data was conducted. After the experiment's conclusion and the process's termination, system-wide CPU and memory utilization data were gathered once more. In order to maintain the integrity and fairness of the experimental results, all experiments were initiated simultaneously. However, due to variations in latency among the different models, each model concluded its execution at different points in time. For these experiments, the PSUTIL package was utilized \cite{Psutil}. PSUTIL is a Python cross-platform library that provides an interface to retrieve information on system utilization, resources, and processes.The size of the BERT large TensorFlow, MobileBERT TensorFlow, non-quantized TensorFlow-Lite MobileBERT, quantized 16 and 8-bit TensorFlow-Lite MobileBERT models are 4GB, 299MB, 98MB, 49MB, and 25MB respectively. For the purpose of experimentation on Linux-based embedded devices, Raspberry Pi 3B, 3B+, and 4B devices \cite{raspberrypi} were utilized. The specifications of the hardware used in this work are presented in Table \ref{tab:hwspec}. The BERT Large and MobileBERT TensorFlow models were deployed on a desktop workstation, while the 8-bit quantized, 16-bit quantized, and non-quantized (32-bit) versions of the MobileBERT TensorFlow-Lite models were deployed on Raspberry Pi devices. This resulted in a total of eleven models for comparative analysis. To facilitate the reproducibility of the results, the same random seeds were utilized for experiments on each device.

\subsubsection{Current Sensing using ACS712 Sensor}

\begin{figure}[!t]
    \centering
    \captionsetup{justification=centering}
    \includegraphics[width=0.9\columnwidth]{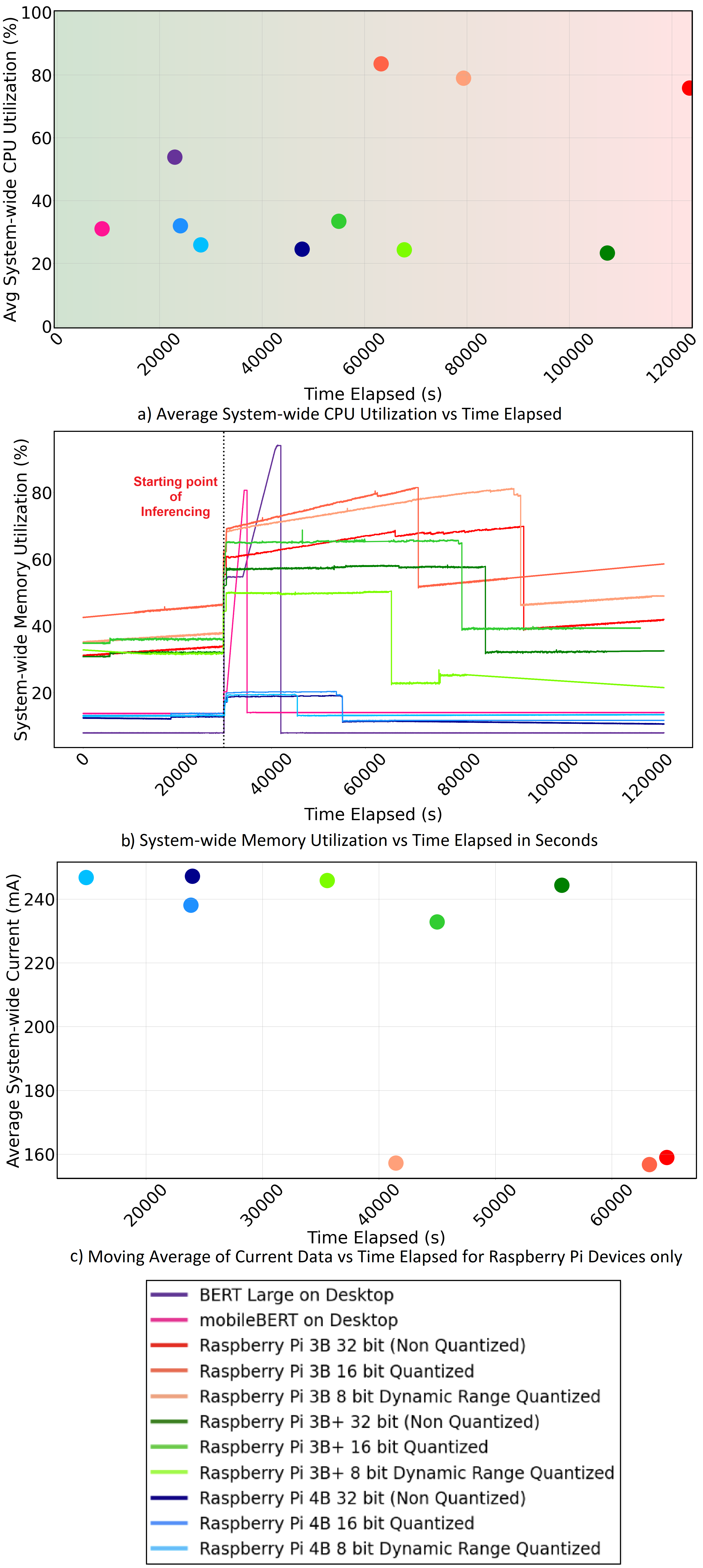}
    \vspace{-0.5em}
    \caption{CPU, memory, and power utilization of the deployed models.}
    \label{fig:resource-usage}
\vspace{-2em}
\end{figure}

The power dissipation of Raspberry Pi modules can be accurately estimated using an external current sensor, such as the ACS712 \cite{ACS712}. This sensor operates based on the Hall effect principle to measure the current flowing through a circuit by detecting the generated Hall voltage. Fig. \ref{current sensing} illustrates the schematic diagram of the circuitry used for current measurement.
By connecting the ACS712 current sensor in series with the load, the sensor can measure the analog hall voltage magnitude corresponding to the instantaneous current. To convert this analog magnitude into a digital format, an Arduino Uno board \cite{ArduinoUno} with a 10-bit analog-to-digital converter (ADC) is used. The formula for calculating the instantaneous current $I(t)$ is:
\begin{equation}
I(t) = \frac{V_h \times V_{ref}}{ADC_{resolution}},
\end{equation}
where $V_h$ represents the instantaneous current sensor reading (Hall Voltage), $V_{ref}$ is the reference voltage used by Arduino Uno (5V), and $ADC_{resolution}$ refers to the resolution of the ADC, which is 10 bits (resulting in 1024 possible values).

\subsection{Results from Comparative Analysis}

\begin{figure}[!t]
    \centering
    \captionsetup{justification=centering}
    \includegraphics[width=0.9\columnwidth]{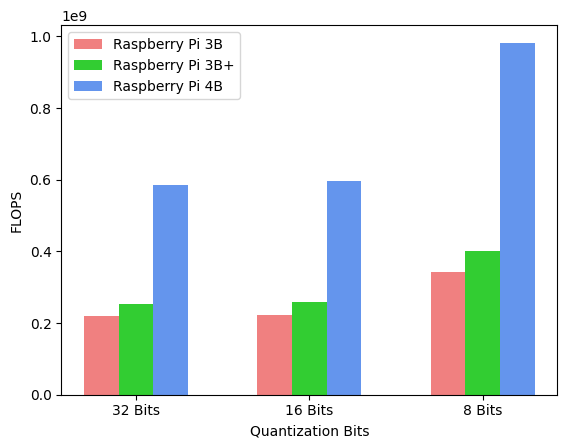}
    \vspace{-1em}
    \caption{FLOPS Comparison for Raspberry Pi Models.}
    \label{fig:FLOPS}
    \vspace{-1.5em}
\end{figure}

\begin{figure}[!t]
    \centering
    \captionsetup{justification=centering}
    \includegraphics[width=0.9\columnwidth]{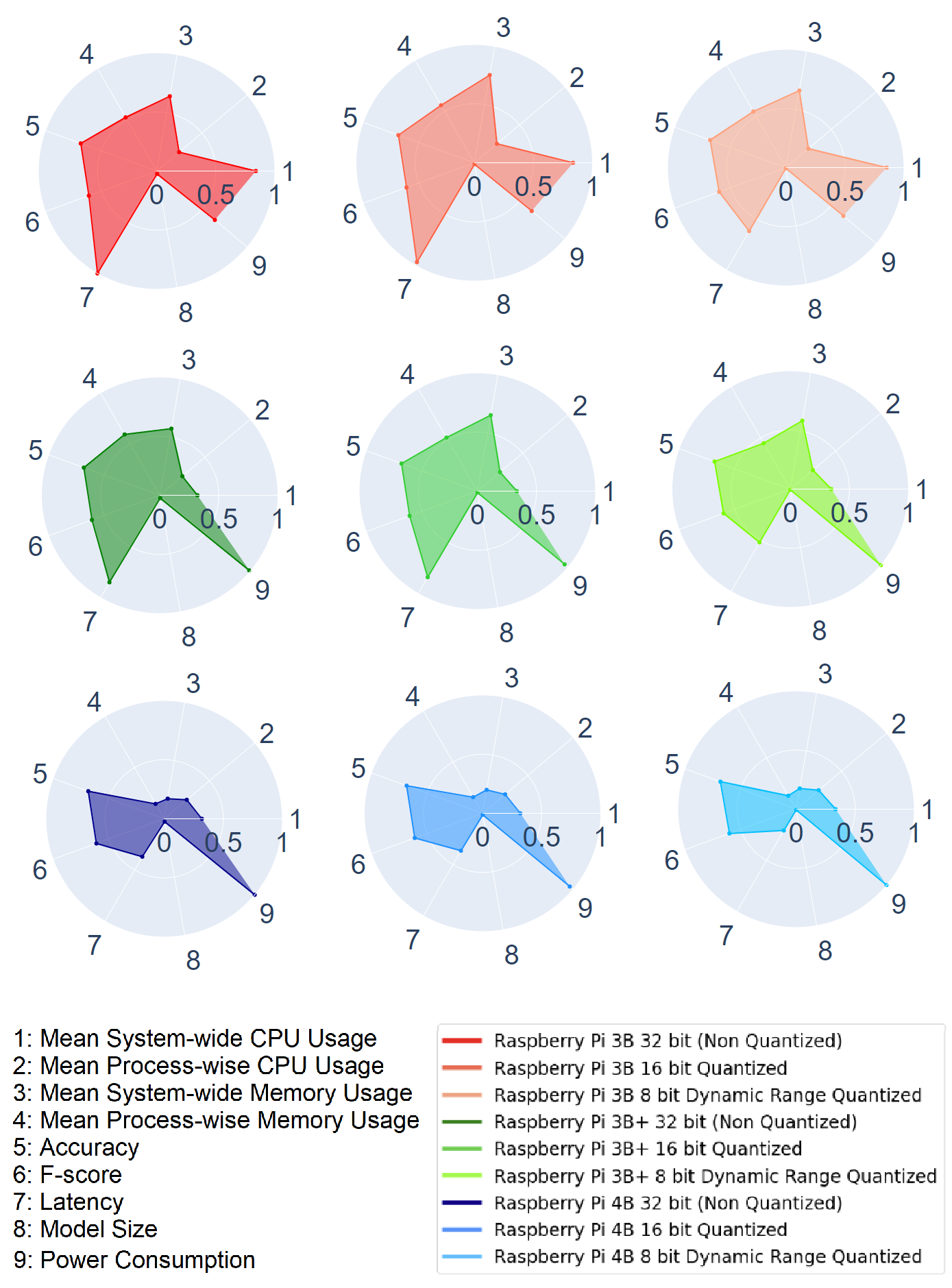}
    \vspace{-0.5em}
    \caption{Comparative Analysis of the deployed models.}
    \label{fig:comparison}
    \vspace{-2em}
\end{figure}

\begin{table*}[!t]
    \centering
    \caption{Model Evaluation Results.}
    \label{tab:evaluation_results}
    \vspace{-0.5em}
     \resizebox{0.9\textwidth}{!}{%
   \begin{tabular}{cccccccccccc}
    \hline\hline
    Device & \thead{Quantization \\ Bits} & \thead{Power \\ (KW)} & \thead{Time\\(s)} & \thead{Power per\\ Inference (W/sample)} & \thead{Time per \\ Inference (s/sample)} & FLOPS & Avg Accuracy  & Avg F-Score  & \thead{Avg CPU \\ Util. (\%)} & \thead{Avg Memory \\ Util. (\%)} \\
    \hline
    RP3B & 32 & 2919.80 & 64719 & 74.9 & 1.66 & $2.18\cdot{10^{8}}$ & 0.685  & 0.602  & 83.9 & 64.1 \\
    RP3B & 16 & 2845.32 & 63219 & 72.99 & 1.62 & $2.23\cdot{10^{8}}$ & 0.683  & 0.601 & 83.5 & 75.3 \\
    RP3B & 8 & 1851.31 & 41449 & 47.49 & 1.06 & $3.42\cdot{10^{8}}$ & 0.684  & 0.603 & 85.2 & 66.1 \\\hline
    RP3B+ & 32 & 3848.44 & 55693 & 98.72 & 1.42 & $2.54\cdot{10^{8}}$ & 0.685  & 0.602 & 31.9 & 57.1 \\
    RP3B+ & 16 & 3651.06 & 44999 & 93.66 & 1.15 & $2.57\cdot{10^{8}}$ & 0.683  & 0.601 & 33.4 & 65.2 \\
    RP3B+ & 8 & 2471.69 & 35549 & 63.4 & 0.91 & $4.00\cdot{10^{8}}$ & 0.684  & 0.603 & 34.7 & 58.7 \\\hline
    RP4B & 32 & 1675.51 & 23969 & 42.98 & 0.61 & $5.84\cdot{10^{8}}$ & 0.685  & 0.602 & 31.9 & 17.2 \\
    RP4B & 16 & 1653.73 & 23849 & 42.42 & 0.6 & $5.97\cdot{10^{8}}$ & 0.683  & 0.601 & 31.9 & 20.1 \\
    RP4B & 8 & 1052.62 & 14851 & 27 & 0.38 & $9.82\cdot{10^{8}}$ & 0.684  & 0.603 & 33.4 & 17.8 \\
    \hline\hline
    \end{tabular}}
    \vspace{-1.5em}
\end{table*}

\begin{figure*}[!t]
    \centering
    \captionsetup{justification=centering}
    \includegraphics[width=1\textwidth]{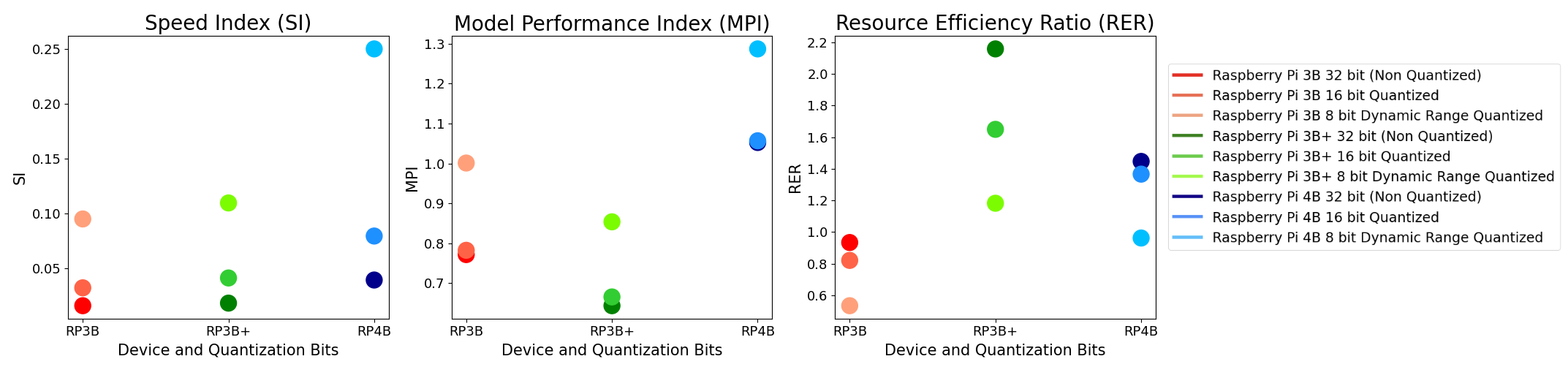}
    \vspace{-1.8em}
    \caption{The SI, MPI, and RER values of the deployed models}
    \label{fig:equation_values}
\vspace{-2em}
\end{figure*}

The quantitative results of our experiments are provided in Table \ref{tab:results}. The results show that the conversion of TensorFlow-Lite models resulted in a relatively small (4.1\%) drop in performance. The main distinction between the models can be observed in their resource utilization. The TensorFlow-Lite models, particularly the 8-bit quantized versions, exhibit significantly lower resource usage, as shown in Figure \ref{fig:resource-usage}(a).  
In the context of Raspberry Pi devices, it has been observed that TensorFlow Lite models exhibit an average CPU utilization of approximately 25\% across various Raspberry Pi versions. This utilization corresponds to the utilization of a single core out of the available four cores on these devices. The BERT Large model running on the Desktop workstation has an average process-wise CPU utilization of 71.8\%. Figure \ref{fig:resource-usage}(a) indicates that the TensorFlow-Lite models deployed on the Raspberry Pi 3B showed higher system-wide CPU usage with high latency as well. This observation can be attributed to the delay added for accessing data from memory sources due to paging. Additionally, Figure \ref{fig:resource-usage}(b) demonstrates that the memory utilization of both BERT Large and MobileBERT models is significantly higher in comparison to the TensorFlow-Lite models.

The evaluation results of the nine models deployed on edge devices are presented in Figure \ref{fig:comparison}, addressing the capabilities and limitations of each model by presenting normalized values between 0 and 1 for a set of attributes. The power dissipation data were collected for the nine TensorFlow-lite models only. According to the figure, the latency of the TensorFlow-Lite models is the only area of concern. Specifically, the non-quantized 32-bit versions of TensorFlow-Lite models on Raspberry Pi 3B and 3B+ are slower than the BERT Large model by a factor of 5.55$\times$ and 4.76$\times$, respectively. However, all of the 8-bit quantized TensorFlow-Lite models on the three devices manage to produce at least one prediction per second. The fastest quantized TensorFlow-Lite model, deployed on Raspberry Pi 4B, is only 1.15$\times$ slower than the BERT Large model while offering 160$\times$ smaller footprint. This demonstrates that it is feasible to deploy large NLP models like BERT variants on edge devices and achieve comparable performance and latency efficiently using the TensorFlow-Lite models.

The presented scatter plot in \ref{fig:resource-usage}(c) and spider graphs in \ref{fig:comparison} reveals notable power dissipation disparities between the Raspberry Pi 3B and the Raspberry Pi 3B+ as well as Raspberry Pi 4B. Specifically, it is evident that the latter two models exhibit elevated power dissipation levels in comparison to the Raspberry Pi 3B. This discrepancy may be attributed to the increased floating-point operations per second (FLOPS) and random-access memory (RAM) capacities inherent in the Raspberry Pi 4B and 3B+ models, which contribute to relatively higher power requirements during operation.
Table \ref{tab:evaluation_results} and Figure \ref{fig:equation_values} present evaluation results for different devices and quantization bits, including Raspberry Pi 3B, 3B+, and 4B. power dissipation and inference time were measured, with Raspberry Pi 3B devices generally exhibiting lower power requirements but longer inference times compared to Raspberry Pi 3B+ and Raspberry Pi 4B. FLOPS as presented in Figure \ref{fig:FLOPS} , a measure of computational performance, was highest for Raspberry Pi 4B across all quantization options. The devices' average accuracy and F-score were comparable, with slight variations based on quantization bits. Resource utilization showed variations, with RP3B+ demonstrating lower average CPU utilization and Raspberry Pi 4B having the lowest average memory utilization.

A higher SI value indicates faster processing speed and higher computational efficiency, as the model is able to perform a larger number of operations (FLOPS) relative to the number of quantization bits and time. Conversely, a lower SI value suggests slower performance and potentially less efficient resource utilization. A higher MPI value indicates better overall performance, reflecting a balance between accuracy and energy consumption. By considering the average accuracy, average F-Score, and power dissipation, the MPI provides a comprehensive assessment of the model's performance. A higher RER value, on the other hand, indicates more efficient utilization of resources, reflecting a better balance between resource consumption and performance.
The analysis of the results reveals that the Raspberry Pi 4B equipped with 8-bit quantization exhibited superior values for both SI and MPI metrics. This outcome suggests that the device achieved higher levels of speed and performance compared to other configurations. Conversely, the Raspberry Pi 3B Plus models displayed remarkably high values for the RER metric across the tested configurations, indicating commendable resource efficiency. These findings hold significant implications for guiding the selection of appropriate devices and quantization configurations based on the desired trade-offs including speed, performance, and resource utilization. 
Considering these metrics, researchers and developers can make informed decisions to strike a balance between the aforementioned factors and meet their specific requirements.




\section{Conclusion}

With TinyML, intelligent decisions can be made on edge devices such as smart home appliances, sensors, and wearables. In this paper, we have explored the application of TinyML in NLP through the fine-tuning of BERT models. Our experiments have used TensorFlow-Lite models to identify the reputation polarity from a given text, demonstrating the potential of these models in enabling automation and intelligence on edge devices. Previous works in TinyML have demonstrated the application of BERT variants on Android devices, performing various local NLP tasks without the need for a server. However, more research is needed regarding the deployment of large NLP models on devices with even fewer resources, such as Raspberry Pi. Our paper contributes to this field by providing a thorough evaluation of the capabilities and limitations of MobileBERT TensorFlow-Lite models deployed on Raspberry Pi devices. The results of our experiments demonstrate that these converted and quantized TensorFlow-Lite models can achieve performance comparable to that of the BERT Large model, with significantly lower resource utilization and a smaller code footprint. Our findings provide valuable insight into the deployment of large NLP models on embedded systems using the concepts of TinyML. Our future work will address the integration of TinyML in the context of federated learning, which presents a promising opportunity by enabling the training of ML models on resource-constrained edge devices while preserving privacy. 

\section*{Acknowledgment}
This work is supported by National Science Foundation (NSF) projects 1624668, 1921485, 2213634, and 2335046 the Department of Energy- National Nuclear Security Administration under Award Number DE-NA0003946, the AGILITY project 4263090, sponsored by Korea Institute for Advancement of Technology (KIAT South Korea), and the University of Arizona's Research, Innovation \& Impact (RII) award for the 'Future Factory'.
 
\bibliographystyle{abbrv}
\bibliography{references}

\begin{thebibliography}{10}

\bibitem{ACS712}
Acs712: Fully integrated, hall-effect-based linear current sensor ic with 2.1
  kvrms voltage isolation and a low-resistance current conductor, available at:
  https://www.allegromicro.com/en/products/sense/current-sensor-ics/zero-to-fifty-amp-integrated-conductor-sensor-ics/acs712.

\bibitem{ArduinoUno}
Arduino uno rev3, available at:
  https://store.arduino.cc/products/arduino-uno-rev3.

\bibitem{tf.lite.Interpreter}
Interpreter interface for running tensorflow lite models., retrieved: January
  2023, available at:
  https://www.tensorflow.org/api\_docs/python/tf/lite/interpreter.

\bibitem{TensorFlowQuantization}
Post-training quantization, retrieved: January 2023, available at:
  https://www.tensorflow.org/lite/performance/.

\bibitem{Psutil}
psutil documentation, retrieved: January 2023, available at:
  https://psutil.readthedocs.io/en/latest/.

\bibitem{raspberrypi}
Raspberry pi products, retrieved: January 2023, available at:
  https://www.raspberrypi.com/products/.

\bibitem{TensorFlowLite}
Tensorflow-lite, retrieved: January 2023,
  https://www.tensorflow.org/lite/guide.

\bibitem{twitterapi}
Twitter api, retrieved: January 2023, available at:
  https://developer.twitter.com/en/docs/twitter-api.

\bibitem{abadi2016tensorflow}
M.~Abadi, P.~Barham, J.~Chen, Z.~Chen, A.~Davis, J.~Dean, M.~Devin,
  S.~Ghemawat, G.~Irving, M.~Isard, et~al.
\newblock Tensorflow: a system for large-scale machine learning.
\newblock In {\em Osdi}, volume~16, pages 265--283. Savannah, GA, USA, 2016.

\bibitem{amigo2013overview}
E.~Amig{\'o}, J.~Carrillo~de Albornoz, I.~Chugur, A.~Corujo, J.~Gonzalo,
  T.~Mart{\'\i}n, E.~Meij, M.~De~Rijke, and D.~Spina.
\newblock Overview of replab 2013: Evaluating online reputation monitoring
  systems.
\newblock In {\em Information Access Evaluation. Multilinguality,
  Multimodality, and Visualization: 4th International Conference of the CLEF
  Initiative, CLEF 2013, Valencia, Spain, September 23-26, 2013. Proceedings
  4}, pages 333--352. Springer, 2013.

\bibitem{devlin2018bert}
J.~Devlin, M.-W. Chang, K.~Lee, and K.~Toutanova.
\newblock Bert: Pre-training of deep bidirectional transformers for language
  understanding.
\newblock {\em arXiv preprint arXiv:1810.04805}, 2018.

\bibitem{han2015deep}
S.~Han, H.~Mao, and W.~J. Dally.
\newblock Deep compression: Compressing deep neural networks with pruning,
  trained quantization and huffman coding.
\newblock {\em arXiv preprint arXiv:1510.00149}, 2015.

\bibitem{he2018amc}
Y.~He, J.~Lin, Z.~Liu, H.~Wang, L.-J. Li, and S.~Han.
\newblock Amc: Automl for model compression and acceleration on mobile devices.
\newblock In {\em Proceedings of the European conference on computer vision
  (ECCV)}, pages 784--800, 2018.

\bibitem{houlsby2019parameter}
N.~Houlsby, A.~Giurgiu, S.~Jastrzebski, B.~Morrone, Q.~De~Laroussilhe,
  A.~Gesmundo, M.~Attariyan, and S.~Gelly.
\newblock Parameter-efficient transfer learning for nlp.
\newblock In {\em International Conference on Machine Learning}, pages
  2790--2799. PMLR, 2019.

\bibitem{iandola2016squeezenet}
F.~N. Iandola, S.~Han, M.~W. Moskewicz, K.~Ashraf, W.~J. Dally, and K.~Keutzer.
\newblock Squeezenet: Alexnet-level accuracy with 50x fewer parameters and< 0.5
  mb model size.
\newblock {\em arXiv preprint arXiv:1602.07360}, 2016.

\bibitem{jacob2018quantization}
B.~Jacob, S.~Kligys, B.~Chen, M.~Zhu, M.~Tang, A.~Howard, H.~Adam, and
  D.~Kalenichenko.
\newblock Quantization and training of neural networks for efficient
  integer-arithmetic-only inference.
\newblock In {\em Proceedings of the IEEE conference on computer vision and
  pattern recognition}, pages 2704--2713, 2018.

\bibitem{jiao2019tinybert}
X.~Jiao, Y.~Yin, L.~Shang, X.~Jiang, X.~Chen, L.~Li, F.~Wang, and Q.~Liu.
\newblock Tinybert: Distilling bert for natural language understanding.
\newblock {\em arXiv preprint arXiv:1909.10351}, 2019.

\bibitem{liu2019roberta}
Y.~Liu, M.~Ott, N.~Goyal, J.~Du, M.~Joshi, D.~Chen, O.~Levy, M.~Lewis,
  L.~Zettlemoyer, and V.~Stoyanov.
\newblock Roberta: A robustly optimized bert pretraining approach.
\newblock {\em arXiv preprint arXiv:1907.11692}, 2019.

\bibitem{marin2022serverless}
E.~Marin, D.~Perino, and R.~Di~Pietro.
\newblock Serverless computing: a security perspective.
\newblock {\em Journal of Cloud Computing}, 11(1):1--12, 2022.

\bibitem{niu2020real}
W.~Niu, Z.~Kong, G.~Yuan, W.~Jiang, J.~Guan, C.~Ding, P.~Zhao, S.~Liu, B.~Ren,
  and Y.~Wang.
\newblock Real-time execution of large-scale language models on mobile.
\newblock {\em arXiv preprint arXiv:2009.06823}, 2020.

\bibitem{niu2021compression}
W.~Niu, Z.~Kong, G.~Yuan, W.~Jiang, J.~Guan, C.~Ding, P.~Zhao, S.~Liu, B.~Ren,
  and Y.~Wang.
\newblock A compression-compilation framework for on-mobile real-time bert
  applications.
\newblock {\em arXiv preprint arXiv:2106.00526}, 2021.

\bibitem{openai2023gpt4}
OpenAI.
\newblock Gpt-4 technical report, 2023.

\bibitem{radford2019language}
A.~Radford, J.~Wu, R.~Child, D.~Luan, D.~Amodei, I.~Sutskever, et~al.
\newblock Language models are unsupervised multitask learners.
\newblock {\em OpenAI blog}, 1(8):9, 2019.

\bibitem{rahman2022bert}
M.~W.~U. Rahman, S.~Shao, P.~Satam, S.~Hariri, C.~Padilla, Z.~Taylor, and
  C.~Nevarez.
\newblock A bert-based deep learning approach for reputation analysis in social
  media.
\newblock In {\em 2022 IEEE/ACS 19th International Conference on Computer
  Systems and Applications (AICCSA)}, pages 1--8. IEEE, 2022.

\bibitem{soro2021tinyubq}
S.~Soro.
\newblock Tinyml for ubiquitous edge ai.
\newblock 2021.

\bibitem{sun2020MobileBERT}
Z.~Sun, H.~Yu, X.~Song, R.~Liu, Y.~Yang, and D.~Zhou.
\newblock Mobilebert: a compact task-agnostic bert for resource-limited
  devices.
\newblock {\em arXiv preprint arXiv:2004.02984}, 2020.

\bibitem{vaswani2017attention}
A.~Vaswani, N.~Shazeer, N.~Parmar, J.~Uszkoreit, L.~Jones, A.~N. Gomez,
  {\L}.~Kaiser, and I.~Polosukhin.
\newblock Attention is all you need.
\newblock {\em Advances in neural information processing systems}, 30, 2017.

\bibitem{wang2020hat}
H.~Wang, Z.~Wu, Z.~Liu, H.~Cai, L.~Zhu, C.~Gan, and S.~Han.
\newblock Hat: Hardware-aware transformers for efficient natural language
  processing.
\newblock {\em arXiv preprint arXiv:2005.14187}, 2020.

\bibitem{wang2019haq}
K.~Wang, Z.~Liu, Y.~Lin, J.~Lin, and S.~Han.
\newblock Haq: Hardware-aware automated quantization with mixed precision.
\newblock In {\em Proceedings of the IEEE/CVF Conference on Computer Vision and
  Pattern Recognition}, pages 8612--8620, 2019.

\bibitem{warden2019tinyml}
P.~Warden and D.~Situnayake.
\newblock {\em Tinyml: Machine learning with tensorflow lite on arduino and
  ultra-low-power microcontrollers}.
\newblock O'Reilly Media, 2019.

\bibitem{wu2020lite}
Z.~Wu, Z.~Liu, J.~Lin, Y.~Lin, and S.~Han.
\newblock Lite transformer with long-short range attention.
\newblock {\em arXiv preprint arXiv:2004.11886}, 2020.

\bibitem{yan2020micronet}
Z.~Yan, H.~Wang, D.~Guo, and S.~Han.
\newblock Micronet for efficient language modeling.
\newblock In {\em NeurIPS 2019 Competition and Demonstration Track}, pages
  215--231. PMLR, 2020.

\bibitem{yang2019xlnet}
Z.~Yang, Z.~Dai, Y.~Yang, J.~Carbonell, R.~R. Salakhutdinov, and Q.~V. Le.
\newblock Xlnet: Generalized autoregressive pretraining for language
  understanding.
\newblock {\em Advances in neural information processing systems}, 32, 2019.

\end{thebibliography}
\end{document}